# A Comparison between Background Modelling Methods for Vehicle Segmentation in Highway Traffic Videos

L. A. Marcomini, A. L. Cunha

**Abstract**

The objective of this paper is to compare the performance of three background-modeling algorithms in segmenting and detecting vehicles in highway traffic videos. All algorithms are available in OpenCV and were all coded in Python. We analyzed seven videos, totaling 2 hours of recording. To compare the algorithms, we created 35 ground-truth images, five from each video, and we used three different metrics: accuracy rate, precision rate, and processing time. By using accuracy and precision, we aim to identify how well the algorithms perform in detection and segmentation, while using the processing time to evaluate the impact on the computational system. Results indicate that all three algorithms had more than 90% of precision rate, while obtaining an average of 80% on accuracy. The algorithm with the lowest impact on processing time allowed the computation of 60 frames per second.

## 1 Introduction

Traditional methods of data acquisition in Transport Engineering requires fieldworks, be it manual or automatic. In manual data acquisition, vehicles or pedestrians, which are usually the objects of interest, have their data collected in real time, with in loco observations or video recordings, and stored for posterior analysis. This manual method is subjected to the natural limitations of human beings. In [1], the authors relate data reliability to factors such as motivation and attention span during fieldwork. In [2], the official Brazilian manual for traffic data acquisition warns that a human observer may be able to reliably count up to 350 vehicles/hour in each traffic direction.

According to [3], there are three categories of sensors for automatic data acquisition: piezoelectric, pneumatic or magnetic sensors, which require cuts on the pavement for their installation, disrupting traffic flow; mobile sensors, installed on the cars, as GPS; and remote sensors, installed above or at the shoulders of traffic lanes, based on radar, sound waves, infrared lights or camera images.

The use of image processing techniques, coupled with computer vision, is becoming common ground. Since the 1980s decade, several authors have been applying and modifying tracking algorithms, usually designed for generic situations, in order to solve specific difficulties related to tracking vehicles and pedestrians in a real-world scenario, with lighting changes and occlusions. In [4], the authors propose a system capable to detect lanes automatically and retrieve simple traffic parameters. In [5], the authors tried to tackle the occlusion issue by using a shape estimation step on their method. In [6], the proposed system is capable to process videos in real-time and to extract vehicles counts and velocity.

Recently, the increase in processing power of personal computers permitted the creation systems with greater complexity. In [7], the authors organized all collected data from videos in a database, with traffic information. In [8], background modeling with lighting changes was tackled with the use of an adaptive background update module coupled with vehicle segmentation based on motion histograms. In [9], the system is able to extract common tracking information, such as counting and velocities, and to classify vehicles in different categories. With the popularization of unmanned aerial vehicles (UAVs), solutions based on a top-down recording perspective are also becoming trivial, as it is possible to see in [10] and [11].

All mentioned papers use tracking methods that apply background modeling as their first step to segment vehicles from the background. In [12], the authors make a detailed theoretical analysis of different background modeling methods, divided into 17 categories and 40 subcategories, outlining capabilities and difficulties of each method. In such a vast group of techniques, choosing the best option is sometimes challenging.

This paper has the objective to evaluate three background-modeling methods available in OpenCV, a free and open computer vision library. The analyzed methods are GMG, MOG, and MOG2, in their respective Python implementation. Python was chosen because it is considered an easy-to-learn programming language, with codes that are easy to understand, as seen on [13] and [14]. It is also indicated for academic purposes by [15].

**2 Background Modelling**

Every frame of a video can be divided into two different areas: foreground, which groups pixels that are part of the objects of interest (in this case, pedestrians, vehicles, or cyclists); and background, which groups all pixels that are not part of an object of interest, such as pavement, trees, sky, buildings [16]. Based on this division, any system with the objective to detect objects automatically, with segmentation and tracking, must be able to differentiate between these two areas [17].

The official OpenCV 3.2 documentation, available on [18], has specific tutorials to explain how to use three background-modeling algorithms: GMG, MOG, and MOG2.

**2.1 GMG**

The GMG algorithm, proposed on [19], models the background with a combination of Bayesian Inference and Kalman Filters. The first stage of the method accumulates, for each pixel, weighted values depending on how long a color stays on that position. For every frame, new observations are added to the model, updating these values. Colors that stay static for a determined amount of time are considered background. The second stage filters pixels on the foreground to reduce noise from the first stage.

The Python implementation of the GMG method has input parameters that may be modified. The default values are presented in Figure 1.

```
cv2.bgsegm.createBackgroundSubtractorGMG(
                                        initializationFrames=120,
                                        decisionThreshold=0.8
                                       )
```

**Figure 1 – GMG constructor, with default values.**

The parameter `initializationFrames` indicates how many frames the algorithm is going to use to initialize the background-modeling method. During the initialization, the resulting frame is always black. The more frames used on this phase, the more stabilized the initial model is. The parameter `decisionThreshold` determines the threshold in which pixels are classified as background or foreground. In the first stage, when the algorithm accumulates values based on the time a color remains static, every pixel with a lower weighted value then the threshold is considered part of the background. Choosing high values for this parameter may result in loss of object detections.

## 2.2 Mixture of Gaussians (MOG)

Mixture of Gaussians, or MOG, was initially proposed on [20], based on [21]. On this method, a mixture of k Gaussians distributions models each background pixel, with values for k within 3 and 5. The authors assume that different distributions represent each different background and foreground colors. The weight of each one of those used distributions on the model is proportional to the amount of time each color stays on that pixel. Therefore, when the weight of a pixel distribution is low, that pixel is classified as foreground.

On OpenCV, the MOG implementation has input parameters that calibrate the behavior of the method. These parameters and their default values may be observed in Figure 2.

```
cv2.bgsegm.createBackgroundSubtractorMOG(
                                         history=200,
                                         nmixtures=5,
                                         backgroundRatio=0.7,
                                         noiseSigma=0
                                        )
```

**Figure 1 – MOG constructor, with default values.**

The parameter `history` is responsible for the number of frames the method will use to accumulate weights on the model, throughout the entire processing period. Low values result in increased sensitivity to sudden changes of luminosity. The parameter `nmixtures` indicates the method how many Gaussians distributions it should during the whole video. Higher values drastically increase processing time. The parameter `backgroundRatio` defines the threshold weight for the differentiation between foreground and background. Lower values may incur in false objects. Finally, the parameter `noiseSigma` defines the accepted noise level. Low values create false objects.

## 2.3 Mixture of Gaussians 2 (MOG2)

The MOG2 method was based on the works of [22] and [23] with the objective to solve one of the limitations that MOG had: the fixed amount of used distributions. By using a variable amount of Gaussians distributions, mapped pixel by pixel, MOG2 achieves a better representation of the complexity of colors in each frame.

On its OpenCV implementation, MOG2 has three input parameters that may be changed to calibrate for each different video. The parameters and their default values may be seen on Figure 3.

```
cv2.createBackgroundSubtractorMOG2(
                                    history=200,
                                    varThreshold=16,
                                    detectShadows=True
                                )
```

**Figure 3 – MOG2 constructor, with default values.**

The `history` parameter functionality is analogue to the first MOG parameter. It denotes the number of frames to be used to model the background. The parameter `varThreshold` correlates the value of the weight of the pixels on the current frame with values on the model. Lower values on this parameter tend to create false objects. The parameter `detectShadows` enables or disables shadow detection. Enabling this parameter increases processing times.

## 3 Proposed Method

This paper's objective is to compare the performance of three background-modeling algorithms available in Python 2.7, on OpenCV 3.2, when applied to vehicle segmentation on highways.

### 3.1 Data

In total, we analyzed seven videos, with different luminosity conditions and camera angles. Two videos were recorded in perspective, as it is possible to see in Figure 4. Video (1) has a cloudy weather, with no shadows, recorded by a monitoring station. Video (2) was captured on top of an overpass, on a sunny day with frequent clouds at noon, with small shadows.

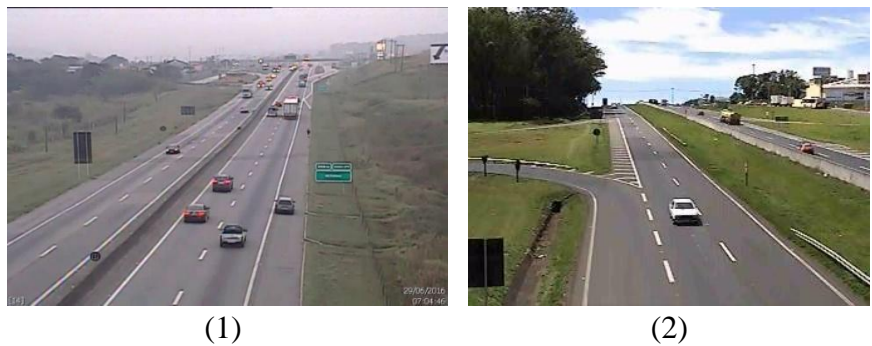

(1)          (2)

**Figure 4 – Videos in perspective.**

Four videos were captured in a frontal perspective with the highway, all on an overpass, as can be seen on Figure 5. Videos (1) and (3) were recorded in a sunny morning, with no clouds and elongated shadows. Video (1) was recorded against the traffic flow and video (3) was recorded in favor of the traffic flow. Video (2) was recorded at noon, on a sunny day and small shadows. Video (4) was captured in a sunny afternoon, with long shadows. Both videos were originated from the work of [24].

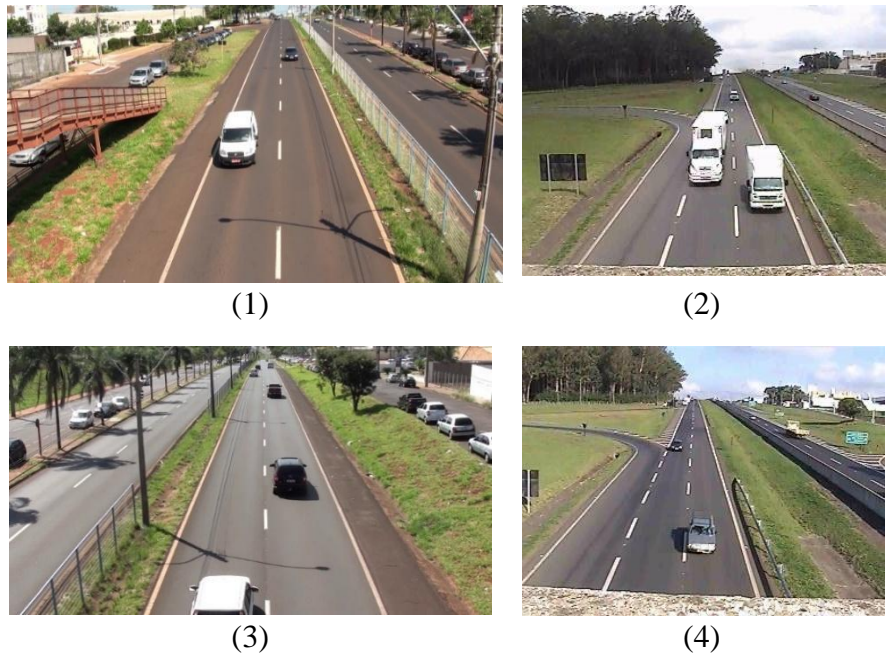

**Figure 5 – Videos with a frontal perspective.**

The final video was recorded in a lateral perspective to the highway, as can be seen in Figure 6. It was captured in the middle of a sunny afternoon, with vehicles driving horizontally on the frame.

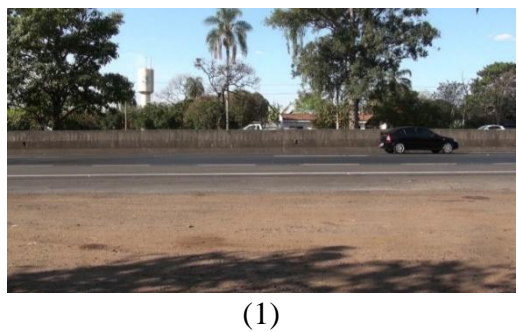

**Figure 6 – Video in lateral perspective.**

### 3.2 Ground-Truth

To compare the methods, we created a database with ground-truth images of each analyzed video. An image with all objects properly segmented, i.e., where all pixels belonging to the foreground are correctly marked, is a ground-truth image. To build our database, we classified manually each image, marking moving vehicles as foreground and everything else as background, including trees, pavement, sky, and parked vehicles.

For each video, five ground-truth images were extracted, with all objects classified accordingly. In total, we created 35 images, distributed along the whole duration of all videos, with the intent to include the largest variety of luminosity conditions available. Examples of ground-truth images extracted can be seen in Figure 7.

**Original Frame**        *Ground-truth*

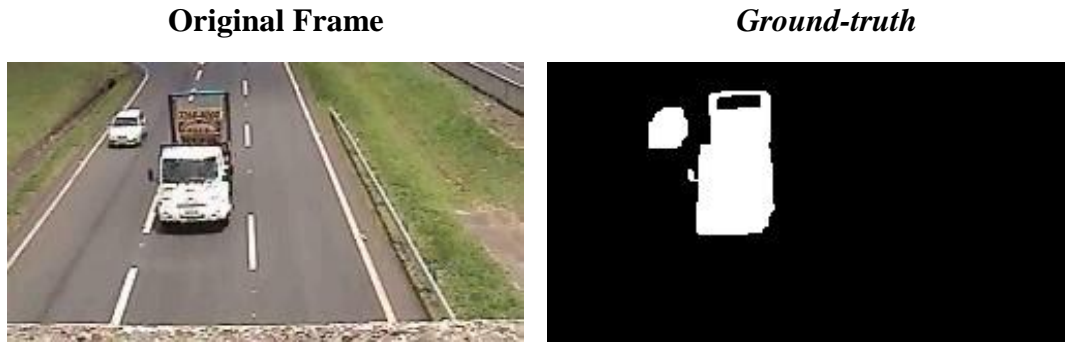

**Figure 7 – Original frame and manually created *ground-truth*.**

**3.3 Background-Modelling Algorithms**

With the manual creation of all ground-truth images, we needed to extract the corresponding processed frame of each tested algorithm. All algorithms use a number of initial frames to stabilize the generated model. Because of that, we cannot start the computation from the wanted frame number. As each algorithm uses a different number of frames, we decided to start all calculation 1000 frames before the frame of interest. For example, to save the 10000th segmented frame of the video, we started the algorithm on frame 9000.

We used the default values of all parameters of the tested algorithms, with no changes. In total, the algorithms created 105 images, 35 each.

It is possible to see examples of output frames from each algorithm in Figure 8.

**Original Frame**        **GMG**

**MOG**        **MOG2**

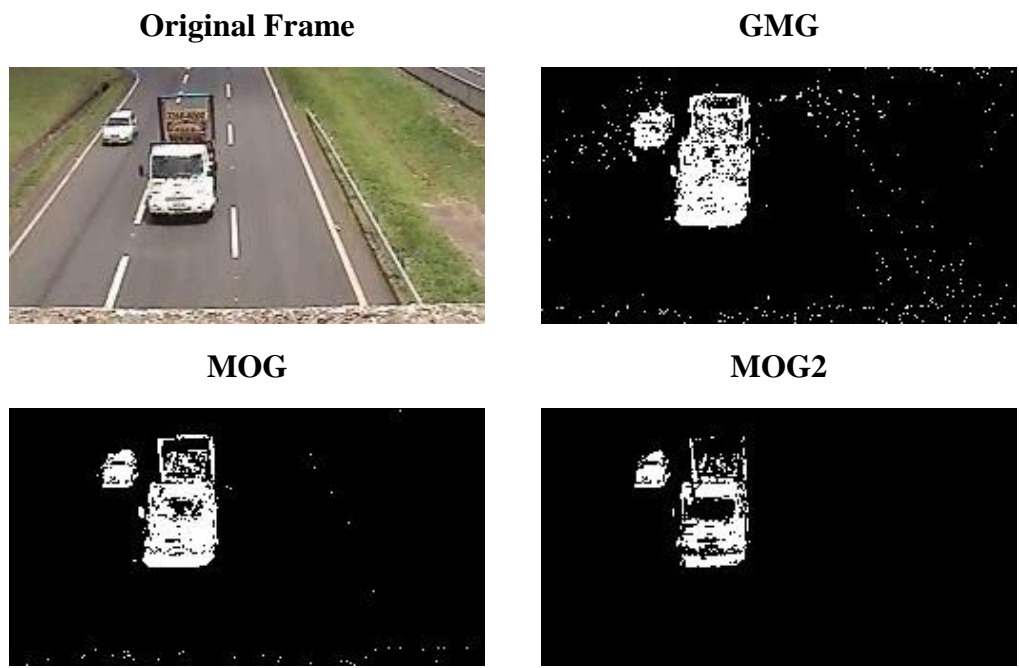

**Figure 8 – Example output frames for each algorithm.**

## 3.4 Evaluation Metrics

The chosen evaluation method follows the recommendations of previous work [25], with the creation of a confusion matrix. The matrix is built according to Table 1.

Table 1 – Confusion Matrix and utilized metrics.

| Model | Ground-Truth Foreground (Positive) | Ground-Truth Background (Negative) |
|---|---|---|
| Foreground (Positive) | TP | FP |
| Background (Negative) | FN | TN |
| **Total** | P | N |

$$\text{Accuracy} = \frac{VP+VN}{P+N}$$

$$\text{Precision} = \frac{VP}{VP+FP}$$

It is possible to notice that we are always comparing the values of pixels from the ground-truth, in the columns, with values of pixels of each modeled frame, in the rows. When both the model and the ground-truth classify a pixel as foreground, we consider it a True Positive (TP). When the model classifies a pixel as foreground, but the ground-truth pixel is a background, we have a False Positive (FP). If the model classifies a pixel as background, but the ground-truth pixel is a foreground, we have a False Negative (FN). Lastly, when both the model and the ground-truth classify a pixel as background, we have a True Negative (TN).

From the confusion matrix, we are able to extract two important metrics: accuracy and precision. Accuracy shows how many pixels were correctly classified in total, taking into consideration both the foreground and the background. Precision, on its turn, denotes only how many foreground pixels the model classified correctly of the total of all foreground pixels the ground-truth had.

Another metric we evaluated was the processing time of each background-modeling algorithm. We obtained the time by using the module `timeit` [26], available natively on Python. With this module, we were able to isolate the specific function call of each algorithm and repeat it several times, allowing us to measure the average time spent at each operation. To measure the time spent, we allowed each algorithm to stabilize its models by giving them 1000 initialization frames. After the stabilization, the `timeit` module repeated the background-modeling line of code several times, giving us the average of time spent per operation.

## 4 Results

Since we selected 5 frames for each one of the 7 available videos, each algorithm was analyzed 35 times. To make the data visualization easier to understand, we organized both accuracy and precision into histograms, in which each bar represents the number of occurrences on that error category. The line on the same plot indicates the accumulated value of occurrences. An optimal result contains several occurrences on the right-most bars, close to 100%.

### 4.1 Accuracy

Accuracy results for the three algorithms can be seen in Figure 9.

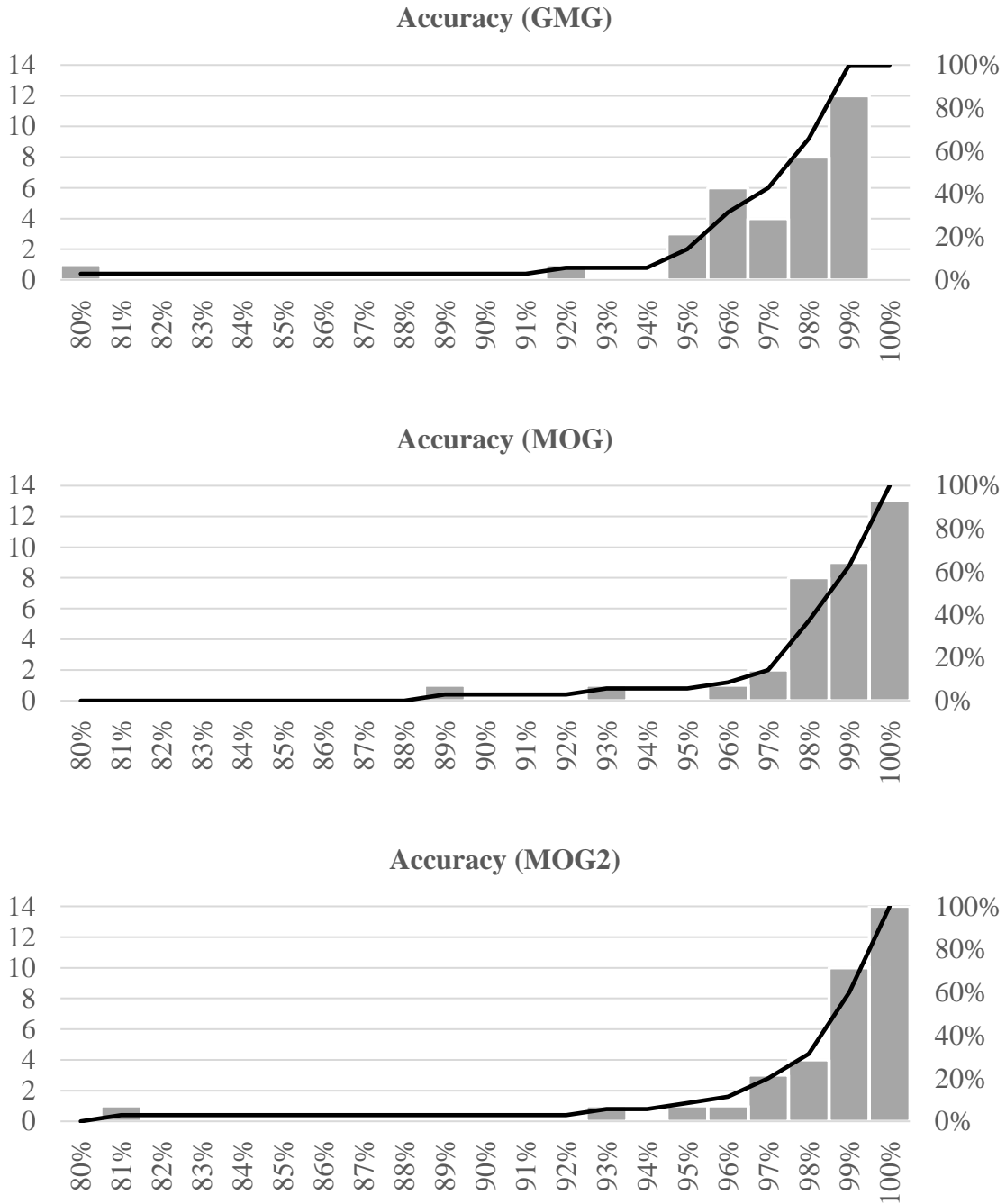

**Figure 9 – Accuracy results for each algorithm.**

It is possible to notice that, for all tested algorithms, results are concentrated above 98%. On GMG, only the video recorded with a lateral perspective resulted in an 80% accuracy rate. All other videos presented rates from 92% to 99% accuracy. On MOG, the lowest accuracy rate observed happened at 89%, with results concentrating above 98%. On MOG2, we can see one result at 81%, also on the lateral video, but results concentrated closer to 100%. Although results show small differences between algorithms, we are not able to select the better option based only on accuracy.

## 4.2 Precision

Results for all three algorithms on precision can be seen in Figure 10.

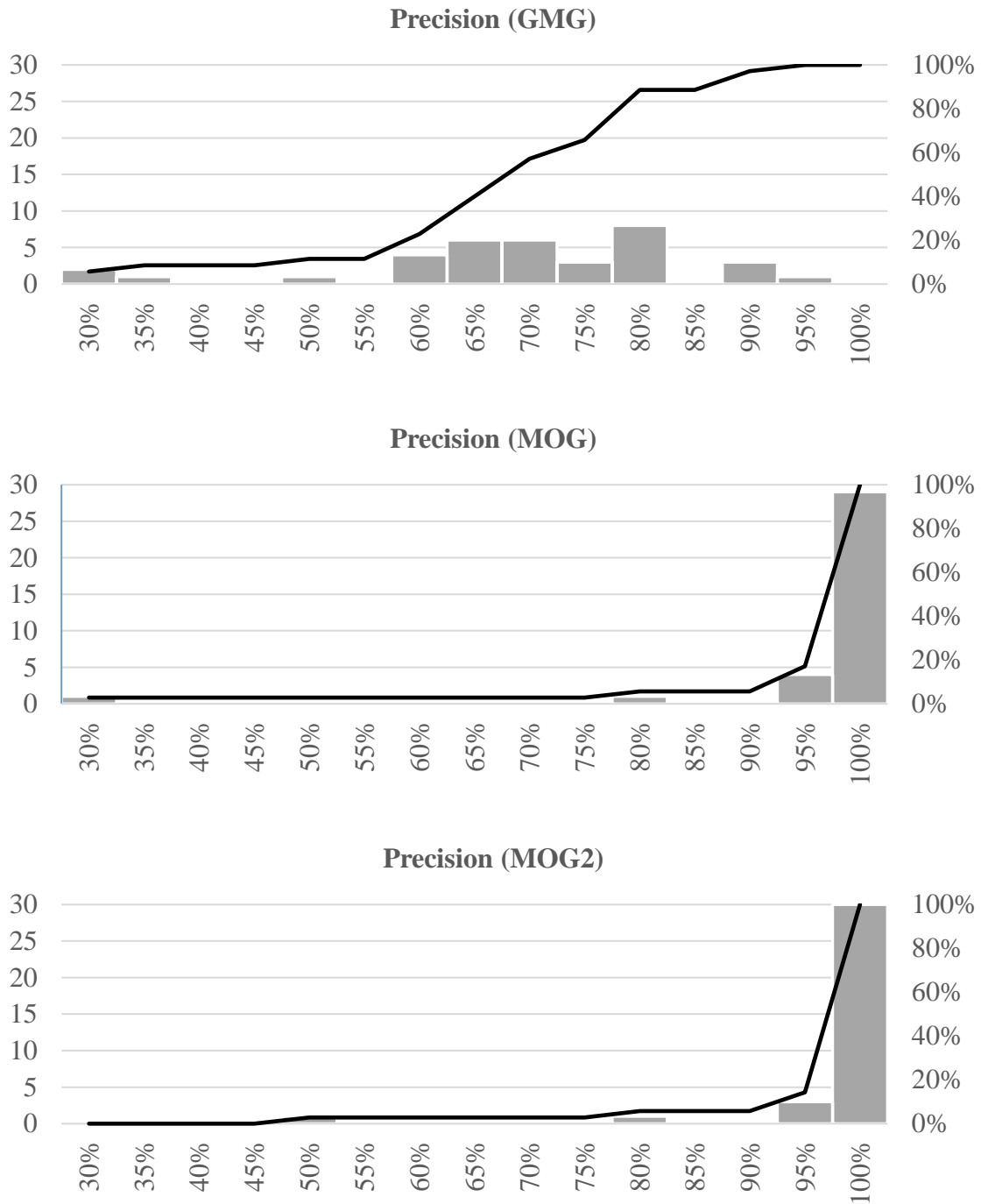

**Figure 10 – Precision results for each algorithm**.

Precision, on the contrary to accuracy, shows performance differences between algorithms. On GMG, results appear distributed along all intervals, with the lowest value at 30%, and concentration of occurrences between 60% and 80%. With MOG and MOG2, results were superior. Both had precision rates near 100%, ie, with pixels correctly classified as foreground. Therefore, based on these results, we can discard GMG as an inferior option for vehicle segmentation, based on our conditions and videos.

### 4.3 Processing Time

To better accrue processing time, tests were executed repeatedly and incrementally. On our first test, each algorithm had to process the same frame 100 times. On our second test, 1,000 times. On our final test, algorithms processed the same frame 10,000 times. On a common scenario, the algorithms apply background-modeling techniques at each frame from the video. Thus, considering a typical video with 30 frames per second, our first test – with 100 repetitions – is similar to processing a video with approximately 3 seconds of duration. Results of our tests can be seen in Figure 11.

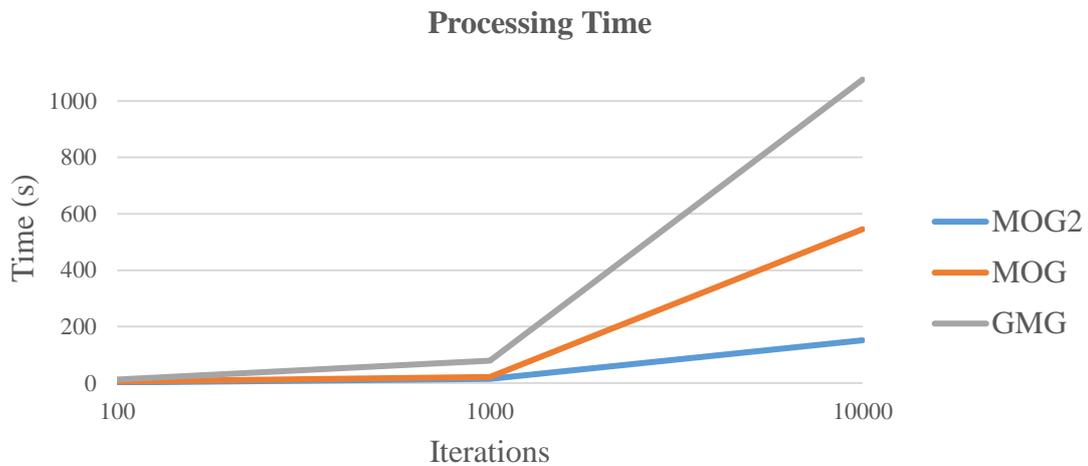

**Figure 11 – Processing time results.**

On average, MOG2 performance results are 3 times better than MOG, and 10 times better then GMG. GMG processed, on average, 10 frames per second and took, on the third test, almost 1,100 seconds to process 10,000 frames. MOG was able to achieve a 28 frames per second rate, processing 10,000 frames on 500 seconds. MOG2, the fastest of the three, processed 10,000 frames on 150 seconds, resulting in a rate of 64 frames per second.

### 5 Conclusions

This paper had the objective to compare three of the available background-modeling algorithms on OpenCV, in Python, and determine which one would be better suited for our needs, based on our videos and conditions, to segment vehicles. We utilized a novel technique, using objective parameters and results to select one algorithm. This technique, as a whole, can be replicated to analyze any number of algorithms, needing only ground-truth images of videos.

As a result, our tests show that MOG2 has a better performance, with accuracy rates equivalent to other tested algorithms, but superior precision rate and lower processing times.